\def\year{2020}\relax
\documentclass[letterpaper]{article} 
\usepackage{aaai20}  
\usepackage{times}  
\usepackage{helvet} 
\usepackage{courier}  
\usepackage[hyphens]{url}  
\usepackage{graphicx} 
\usepackage{graphicx}  
\usepackage{tikz}
\usepackage{xcolor}
\usepackage{booktabs}
\usepackage{mathtools}
\usetikzlibrary{arrows, decorations.markings}
\usetikzlibrary{fit, positioning, shapes, calc, patterns}
\usepackage[utf8]{inputenc}
\usepackage{subcaption}
\usepackage{pgfplots}

\title{Towards Graph Representation Learning in Emergent Communication}
\author{\Large \textbf{Agnieszka S\l{}owik}\thanks{Equal Contribution} \textsuperscript{\rm 1} \Large \textbf{Abhinav Gupta}$^*$ \textsuperscript{\rm 2}\\ \Large \textbf{William L. Hamilton} \textsuperscript{\rm 2}\textsuperscript{\rm 3} \Large \textbf{Mateja Jamnik} \textsuperscript{\rm 1} \Large \textbf{Sean B. Holden} \textsuperscript{\rm 1}\\
\textsuperscript{\rm 1} University of Cambridge  
\textsuperscript{\rm 2} MILA
\textsuperscript{\rm 3} McGill University\\
agnieszka.slowik@cl.cam.ac.uk, abhinavg@nyu.edu
}

\begin{document}

\maketitle

\begin{abstract}
Recent findings in neuroscience suggest that the human brain represents information in a geometric structure (for instance, through conceptual spaces). In order to communicate, we flatten the complex representation of entities and their attributes into a single word or a sentence. In this paper we use graph convolutional networks to support the evolution of language and cooperation in multi-agent systems. Motivated by an image-based referential game, we propose a graph referential game with varying degrees of complexity, and we provide strong baseline models that exhibit desirable properties in terms of language emergence and cooperation. We show that the emerged communication protocol is robust, that the agents uncover the true factors of variation in the game, and that they learn to generalize beyond the samples encountered during training.
\end{abstract}
\section{Introduction}

The ability to represent complex concepts and the relationships between them in the manner of a mental graph was found to be one of the key factors behind knowledge generalization and prolonged learning \cite{Bellmundeaat6766}. Through communication, humans flatten the non-Euclidean representation of ideas into a sequence of words. Advances in graph representation learning \cite{kipf2017semi} and sequence decoding \cite{sutskever_sequence_2014,cho-etal-2014-learning} provide the means for simulating this \textit{graph linearization} process in the multi-agent setting.

One of the approaches to learning to communicate in a multi-agent system is through emergent communication \cite{sukhbaatar2016learning,lazaridou_multi-agent_2016,foerster_learning_2016}. In contrast to training a dialogue system in a supervised way, emergent communication supports development of a grounded, goal-oriented and compositional language \cite{mordatch_emergence_2017,lazaridou_emergence_2018,Resnick2019CapacityBA}. The agents develop a language from scratch in order to solve a task in an end-to-end virtual environment, and this allows an extensive study of the communication protocols. Such studies contribute to the long-standing quest to develop multi-agent systems that support a biologically-inspired evolution of cooperation through language \cite{brennan_conceptual_1996,smith_complex_2004}. 

Existing environments for emergent communication use sequences of one-hot vectors or images as the input data \cite{lazaridou_emergence_2018,evtimova2018emergent,bouchacourt-baroni-2018-agents}. The degree of structure in the training samples was found to affect the evolution of compositional understanding \cite{Smith:2003:ILF:963725.963729,kirby_iterated_2014,raviv_systematicity_2018}, with more structured prelinguistic representations leading to a more compositional language \cite{lazaridou_emergence_2018}. Compositionality is one of the most desired properties in a communication protocol because it allows the agents to understand an (in principle) infinite number of complex structures through a finite vocabulary.

Motivated by the properties of graph representations---namely, their explicit structural bias \cite{hamilton_inductive_2017}, wider array of applications than when using sequences and images, and ability to encode rich compositional properties---we introduce graph representation learning to emergent communication.

\begin{figure}[t]
\noindent
\resizebox{\columnwidth}{!}{
\begin{tikzpicture}
\node[circle,draw,fill=yellow!20,minimum size=18pt,label=below:{Line}] (l1) at (0, 0) {};
\node[circle,draw,fill=green!20,minimum size=18pt,label=below:{Color}] (l2) at (1, 0) {};
\node[circle,draw,fill=red!20,minimum size=18pt,label=below:{Shape}] (l3) at (2, 0) {};
\node[circle,draw,fill=white,minimum size=18pt] (l4) at (1, 1.5) {};
\node[draw,fill=yellow!20,inner sep=8pt] (sender) at (4, 1) {Sender};
\node[draw,fill=yellow!20,inner sep=8pt] (receiver) at (7, 1) {Receiver};
\draw[thick] (l1.north) -- (l4.south west);
\draw[thick] (l2.north) -- (l4.south);
\draw[thick] (l3.north) -- (l4.south east);
\draw[very thick, ->] (2.5, 1) -- (sender.west);
\draw[very thick, ->] (sender.east) -- (receiver.west) node[midway,above] {msg};
\draw[very thick, ->] (receiver.east) -- (8.5, 1);
\node[circle,draw,fill=yellow!20,minimum size=18pt,label=below:{Line}] (r1) at (9, 0) {};
\node[circle,draw,fill=green!20,minimum size=18pt,label=below:{Color}] (r2) at (10, 0) {};
\node[circle,draw,fill=red!20,minimum size=18pt,label=below:{Shape}] (r3) at (11, 0) {};
\node[circle,draw,fill=white,minimum size=18pt] (r4) at (10, 1.5) {};
\draw[thick] (r1.north) -- (r4.south west);
\draw[thick] (r2.north) -- (r4.south);
\draw[thick] (r3.north) -- (r4.south east);


\node[circle,draw,fill=yellow!20,minimum size=18pt,label=below:{Line}] (dr1) at (9, -4) {};
\node[circle,draw,fill=green!20,minimum size=18pt,label=below:{Color}] (dr2) at (10, -4) {};
\node[circle,draw,fill=red!20,minimum size=18pt,label=below:{Shape}] (dr3) at (11, -4) {};
\node[circle,draw,fill=white,minimum size=18pt] (dr4) at (10, -2.5) {};
\draw[thick] (dr1.north) -- (dr4.south west);
\draw[thick] (dr2.north) -- (dr4.south);
\draw[thick] (dr3.north) -- (dr4.south east);

\node[circle,draw,fill=yellow!20,minimum size=18pt,label=below:{Line}] (dl1) at (3, -4) {};
\node[circle,draw,fill=green!20,minimum size=18pt,label=below:{Color}] (dl2) at (4, -4) {};
\node[circle,draw,fill=blue!20,minimum size=18pt,label=below:{Shape}] (dl3) at (5, -4) {};
\node[circle,draw,fill=white,minimum size=18pt] (dl4) at (4, -2.5) {};
\draw[thick] (dl1.north) -- (dl4.south west);
\draw[thick] (dl2.north) -- (dl4.south);
\draw[thick] (dl3.north) -- (dl4.south east);

\node[circle,draw,fill=purple!20,minimum size=18pt,label=below:{Line}] (dc1) at (6, -4) {};
\node[circle,draw,fill=green!20,minimum size=18pt,label=below:{Color}] (dc2) at (7, -4) {};
\node[circle,draw,fill=red!20,minimum size=18pt,label=below:{Shape}] (dc3) at (8, -4) {};
\node[circle,draw,fill=white,minimum size=18pt] (dc4) at (7, -2.5) {};
\draw[thick] (dc1.north) -- (dc4.south west);
\draw[thick] (dc2.north) -- (dc4.south);
\draw[thick] (dc3.north) -- (dc4.south east);

\draw [thick,decoration=brace,decorate] (3,-2) -- (11,-2);

\draw[dashed] (7,-1.5) -- (7,0.5);

\end{tikzpicture}}
\caption{Intuition behind the graph referential game. The colored graph nodes correspond to independent properties (line, color, shape). The receiver identifies the target graph among distractors based on the message transmitted by the sender.}
\label{fig:game}
\end{figure}
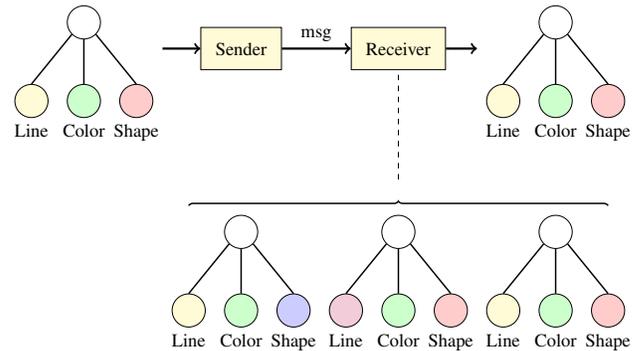

Our contributions are as follows:

\begin{itemize}
\item We propose a graph referential game with varying degrees of complexity, based on the number of distractors, vocabulary size, properties, and the corresponding types present in the graph.
\item We provide baselines using Graph Convolutional Networks \cite{kipf2017semi} and empirically show that these 
models generalize to graphs outside of the training set.
\item We analyze the resulting communication protocol and find that the agents are able to make use of the available symbols in an efficient way.
\item We show that the communication channel is robust in terms of permutation invariance and that it promotes cooperation between the agents.
\end{itemize}

\begin{figure*}[ht]
    \centering
    \includegraphics[width=0.9\linewidth,height=4.5cm]{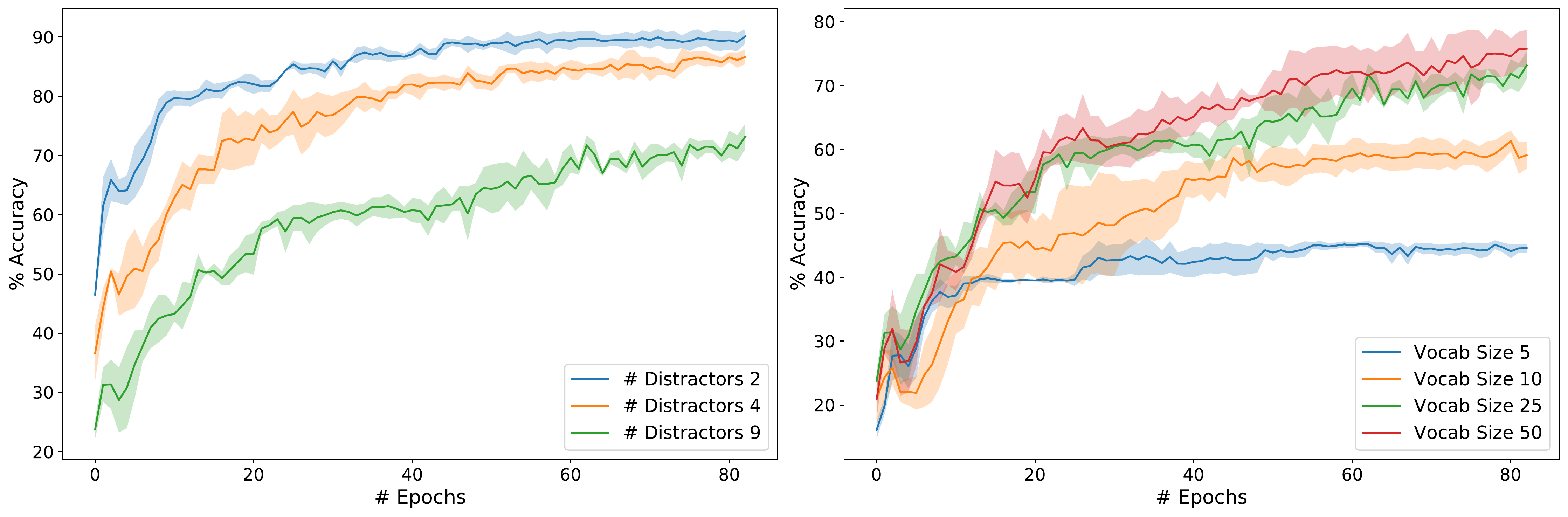}
        \caption{Learning curves showing the performance of the agents on the test set. Left: We vary the number of distractors and fix the vocabulary size (25). Right: We vary the vocabulary size for a fixed number of distractors (4).} 
    \label{fig:learning}
\end{figure*}
\begin{table*}
\centering
{
\small
\begin{tabular}{c|c|c|c}
Vocab Size & 2 distractors & 4 distractors & 9 distractors \\
\toprule
5 & $5.0 \pm 0.0$ ($100 \pm 0$) & $5.0 \pm 0.0$ ($100 \pm 0$) & $5.0 \pm 0.0$ ($100 \pm 0$) \\
10 & $9.33 \pm 0.47$ ($93.33  \pm 4.71$) &  $9.33 \pm 0.47$ ($93.33  \pm 4.71$) & $9.0 \pm 0.81$ ($90 \pm 8.16$) \\
25 & $12.66 \pm 0.47$ ($50.66 \pm 1.88$) & $14.66 \pm 0.47$ ($58.66 \pm 1.88$) & $13.33 \pm 0.47$ ($53.33 \pm 1.88$) \\
50 & $14.0 \pm 1.41$ ($28.0 \pm 2.83$) & $16.33 \pm 0.94$ ($32.66 \pm 1.89$) & $16.66 \pm 0.47$ ($33.33 \pm 0.94$) \\
\end{tabular}
}
\caption{The expected number of symbols the system used per number of distractors. In the parentheses we include the percentage of symbols used for the given vocabulary size. All the values are averaged across three different random seeds and standard errors are shown.}
\label{table:symbol-us}
\end{table*}

\section{Environment}
\label{sec:testtst}
\subsection{Multi Agent Reinforcement Learning}
We define the game with agents deployed in a Multi Agent Reinforcement Learning (MARL) framework. We consider multi-agent Markov games \cite{littman1994markov}. A Markov game for $N$ agents is a partially observable Markov Decision Process (MDP) defined by: a set of states $S$ describing the state of the world and the possible joint configuration of all the agents, a set of observations $O^1,\ldots, O^N$ for each agent, a set of actions for each agent $A^1,\ldots, A^N$, a transition function $T :S \times A^1 \ldots A^N \xrightarrow{} S$ determining a distribution over the next states, and a reward for each agent $i$, which is a function of the state and the agent’s action $r^i: S \times A^i \xrightarrow{} R$. Agents choose their actions according to a stochastic policy $\pi_{\theta^i} : O^i \times  A^i \xrightarrow{} [0, 1]$, where $\theta^i$ are the parameters of the policy. Each agent $i$ aims to maximize its own total expected return $R^i = \sum_{t=0}^{T} \gamma^t r^i_t$, where $\gamma$ is a discount factor and $T$ is the time horizon.

In this paper we refer to the Lewis Signalling Game \cite{lewis1969convention}, which is extensively used in multi-agent communication. This referential game \cite{lazaridou_emergence_2018} involves two agents: the sender and the receiver. The agents communicate with each other in order to learn to distinguish a given \textit{target} among a set of \textit{distractors} (other samples from the target distribution). The sender only `sees' the target and produces a message. The message is received by the receiver along with a set of distractors and the target. The agents are rewarded with a reward $r \in R$ if they correctly identify the target. This framework can be considered as a cooperative partially-observable Markov game. The target and the distractors represent permutations of combinatorial properties represented as symbolic vectors, images, graphs or any efficient data structure able to encode composed entities.

\begin{figure*}[t]
    \centering
    \includegraphics[width=1.0\linewidth]{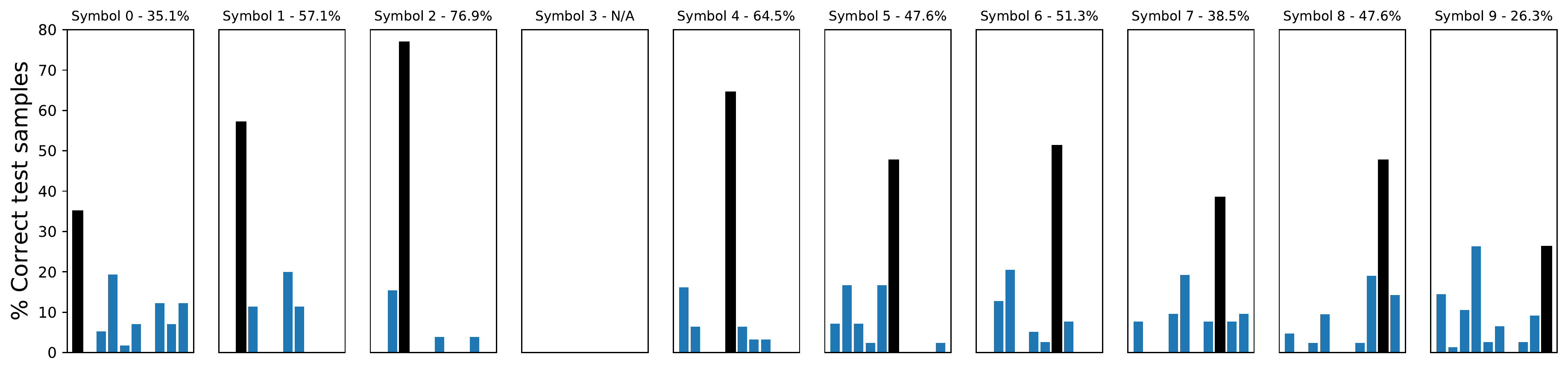}
        \caption{Robustness of the communication protocol learnt by the sender/receiver. We show that for a symbol different than the one sent by the sender (black), the receiver does not correctly identify the target. For more details, refer to Section~\ref{sec:results}. The above chart is for vocab size 10 and number of distractors 4. Note that there were no test samples found where the sender used the symbol $3$.
        }
    \label{fig:dist_4}
\end{figure*}

\subsection{Graph Referential Game}

In our game, we represent the set of properties as unique nodes in a graph. Unlike sequential encoding, this representation allows us to build a hierarchy of concepts, where parent nodes are composed of basic properties encoded in their children. Although this can be achieved using sequences as well, we hypothesize that using graph representations improves the compositional understanding of the agents.

We use $p$ properties of $t$ types, which amounts to $t^p$ unique combinations. As shown in Figure~\ref{fig:game}, each object is represented using a graph $\mathcal{G}(\mathcal{V},\mathcal{E})$ where $\mathcal{V}$ corresponds to the set of all nodes representing unique properties, and a `central' node such that $|\mathcal{V}|=p+1$. The set $\mathcal{E}$ comprises undirected edges, which connect two nodes with a relation. In our simple game, $\mathcal{E}$ consists of the edges between the central node and its children, that represent individual properties, such that $|\mathcal{E}|=p$. All of the nodes except the central node are represented using node features. The node features consist of a concatenation of the property encoding and the type encoding (represented as one-hot vectors). The central node is encoded as an empty node and no edge features are used.

For the purpose of providing a graph baseline, we use one level of concepts. This can be easily extended to a deeper hierarchy of concepts using the aforementioned graph representations. The target and distractors are randomly sampled from the set of all possible graphs without replacement. Each \textit{sample} in the game dataset $\mathcal{D}$ consists of a target graph $d^*$ and the set of $K$ distractors. We obtain a collection of these samples and create the train, validation and test splits (60\%/20\%/20\%).

The sender $f_\theta$ and the receiver $g_\phi$ are parameterized using graph convolutional networks (GCNs) \cite{kipf2017semi}. 
The sender takes the target graph $d^*$ as input and produces a softmax distribution over the vocabulary $V$, where $V$ refers to the finite set of all distinct messages generated by the sender. In this work the message always comprises one word, i.e. a single symbol from $V$. Similar to \cite{sukhbaatar2016learning,mordatch_emergence_2017}, we use the `straight through' version of Gumbel-Softmax \cite{jang_categorical_2016,maddison2016concrete} during training to make the message discrete. At test time, we take the \texttt{argmax} over this distribution.

The receiver takes two inputs: the discretized message $m$ sent by the sender along with the set of distractors $K$ and the target $d^*$. It then outputs a softmax distribution over the $|K|+1$ embeddings representing each graph. We formally define this as follows:

$$
   m(d^*) = \texttt{Gumbel-Softmax}(f_\theta(d^*)) 
$$
$$
    o(m, \{K, d^*\}) = g_\phi(m, \{K, d^*\})
$$
For reference, the graph convolution network is defined as:

$$
     H^{(l+1)} = \sigma(\tilde{D}^{-\frac{1}{2}}\tilde{A}\tilde{D}^{-\frac{1}{2}}H^{(l)}W^{(l)})   
$$

\noindent where $H^{(l)}$ refers to the $l^{th}$ layer in the network, $\sigma$ is the non-linearity, $W$ corresponds to the weight matrix of the $l^{th}$ layer, and $\tilde{D}$ and $\tilde{A}$ represent the normalized degree matrix and adjacency matrix of a graph, respectively. In order to compute the graph embedding, we experimented with the standard graph pooling methods: mean, sum and max functions. We found no significant difference in performance, and thus use mean pooling throughout the experiments presented in this paper. We used the Deep Graph Library \cite{wang2019dgl} when using graphs and EGG \cite{kharitonov-etal-2019-egg} for building the framework while the whole codebase was written using PyTorch \cite{pytorch_NIPS2019}.

\section{Results \& Analysis}
\label{sec:results}

We present the results for the graphs of $p=3$ properties and $t=4$ types. In Figure~\ref{fig:learning}, we confirm that an increase in the number of distractors leads to a higher complexity of the game (measured by the decrease in the performance as the complexity grows). The baseline accuracy for a game with 2, 4 and 9 distractors is 33\%, 20\% and 10\%, respectively, which
corresponds to choosing the target at random. A naive way to solve this game would be to learn unique symbols for each unique graph. For our game, this means that the agents would require at least $4^3=64$ different symbols in the vocabulary. However, we observe that the agents are able to solve the game with a fraction of this vocabulary. We posit that this is due to the agents learning some compositional properties that are encoded in the graph data structure and are essential to solving the game. 

In the right image of Figure~\ref{fig:learning}, we observe that there exists a lower bound on the size of the vocabulary with which the agents are able to achieve high accuracy on the task. This hypothesis is also supported in Table~\ref{table:symbol-us}, which shows the symbol usage across different sizes of the vocabulary. Figure~\ref{fig:learning} (Left) shows that the models achieve the accuracy of $90\%$ for $2$ distractors with less than $25$ symbols. This implies that the sender and the receiver learnt to refer to more than one graph using the same symbol. We hypothesize that graphs mapped to one symbol share a similar structure. Another observation is that the variance of the test accuracies was found to be lower for the higher vocabulary size. We attribute this to the stability in training when there is no pressure in the message channel and higher variance of gradients due to approximate backpropagation (Gumbel-Softmax).

In Figure~\ref{fig:dist_4}, we analyze the robustness of the communication protocols learnt by the sender/receiver. For each symbol $i$ in the vocabulary, we collected the set of correct test samples $D_i$ where the sender used the corresponding symbol to represent the target graph. Each subplot in Figure~\ref{fig:dist_4} represents the distribution of $D_i$ over the whole vocabulary size $|V|=10$. We observe that in all cases the symbol sent by the sender (referred to by the title in each subplot and the position of the bar) is the one that makes the receiver correctly identify the target graph. We think that it shows that the receiver actually cooperates with the sender and uses the message to identify the target among distractors. 

We also analyzed the behavior of the receiver when presented with a shuffled set of graphs. In each experiment, the position of the target is permuted across all of the possible $|K|+1$ positions. We observe that the receiver is still able to correctly identify the target based on the message sent from the sender. We thus posit that the agents learnt an order-invariant representation of the graphs and not some positional information about the ordering of the graphs.

\section{Future Work}

We present results of our work in progress. We are currently designing the set of distractors in a way that allows us to study systematic generalization in our game. One can design the game samples such that the target and the distractors differ in $k$ types, where $k = 1, \dots, p$ and $p$ is the number of properties. In order to study compositionality in the emerged language, we will extend the sender with a sequence decoder and produce messages of multiple symbols. It would be interesting to see if the sender parameterized with a Graph2Seq \cite{xu2019graphseq} network is able to generate sentences close to natural language in terms of compositionality. Another possible direction can be to use a deeper hierarchy of concepts in the graph representations. We expect that hierarchical concepts will show the advantage of using graph representations over sequences and images in multi-agent communication.

\section{Conclusion}

We proposed a new referential game defined on graphs. We showed that agents using simple graph neural networks generalized to new combinations of familiar concepts and types. We found that the agents made an efficient use of the vocabulary, learnt an order-invariant representation of the target graph, and solved the graph games with a varying number of distractors through communication and cooperation.

\bibliography{refs}
\bibliographystyle{aaai}
\end{document}